\documentclass[letterpaper]{article} 
\usepackage{aaai2026}  
\usepackage{times}  
\usepackage{helvet}  
\usepackage{courier}  
\usepackage[hyphens]{url}  
\usepackage{graphicx} 
\usepackage{natbib}  
\usepackage{caption} 
\usepackage{booktabs}
\usepackage{graphicx}
\usepackage{subcaption}

\usepackage{algorithm}
\usepackage{algorithmic}

\usepackage{newfloat}
\usepackage{listings}
\DeclareCaptionStyle{ruled}{labelfont=normalfont,labelsep=colon,strut=off} 
\floatstyle{ruled}
\newfloat{listing}{tb}{lst}{}
\floatname{listing}{Listing}
\nocopyright 

\title{Countering the Over-Reliance Trap: Mitigating Object Hallucination for LVLMs via a Self-Validation Framework}
\author{
    Shiyu Liu$^{1}$\equalcontrib, Xinyi Wen$^{2}$\equalcontrib, Zhibin Lan$^{2}$, Ante Wang$^{2}$,
    Jinsong Su$^{1,2}$\thanks{Corresponding author.}
}
\affiliations{
    \textsuperscript{\rm 1}Institute of Artificial Intelligence, Xiamen University\\
    \textsuperscript{\rm 2}School of Informatics, Xiamen University\\

}

\usepackage{bibentry}
\usepackage{multirow}
\usepackage{amsmath}

\begin{document}


\maketitle

\begin{abstract}
Despite progress in Large Vision Language Models (LVLMs), object hallucination remains a critical issue in image captioning task, where models generate descriptions of non-existent objects, compromising their reliability. 
Previous work attributes this to LVLMs' over-reliance on language priors and attempts to mitigate it through logits calibration. 
However, they still lack a thorough analysis of the over-reliance.
To gain a deeper understanding of over-reliance, we conduct a series of preliminary experiments, indicating that as the generation length increases, LVLMs' over-reliance on language priors leads to inflated probability of hallucinated object tokens, consequently exacerbating object hallucination.
To circumvent this issue, we propose \textbf{Language-Prior-Free Verification} to enable LVLMs to faithfully verify the confidence of object existence. 
%
%
Based on this, we propose a novel training-free \textbf{Self-Validation Framework} to counter the over-reliance trap. It first validates objects' existence in sampled candidate captions and further mitigates object hallucination via caption selection or aggregation. 
Experiment results demonstrate that our framework mitigates object hallucination significantly in image captioning task (e.g., 65.6\% improvement on $\text{CHAIR}_I$ metric with LLaVA-v1.5-7B), surpassing the previous SOTA methods. This result highlights a novel path towards mitigating hallucination by unlocking the inherent potential within LVLMs themselves.\footnote{Code is available at \url{https://github.com/Liushiyu-0709/SelfVal}
.}
\end{abstract}

\begin{figure*}[t] 
    \centering
    \includegraphics[width=1.0\linewidth]{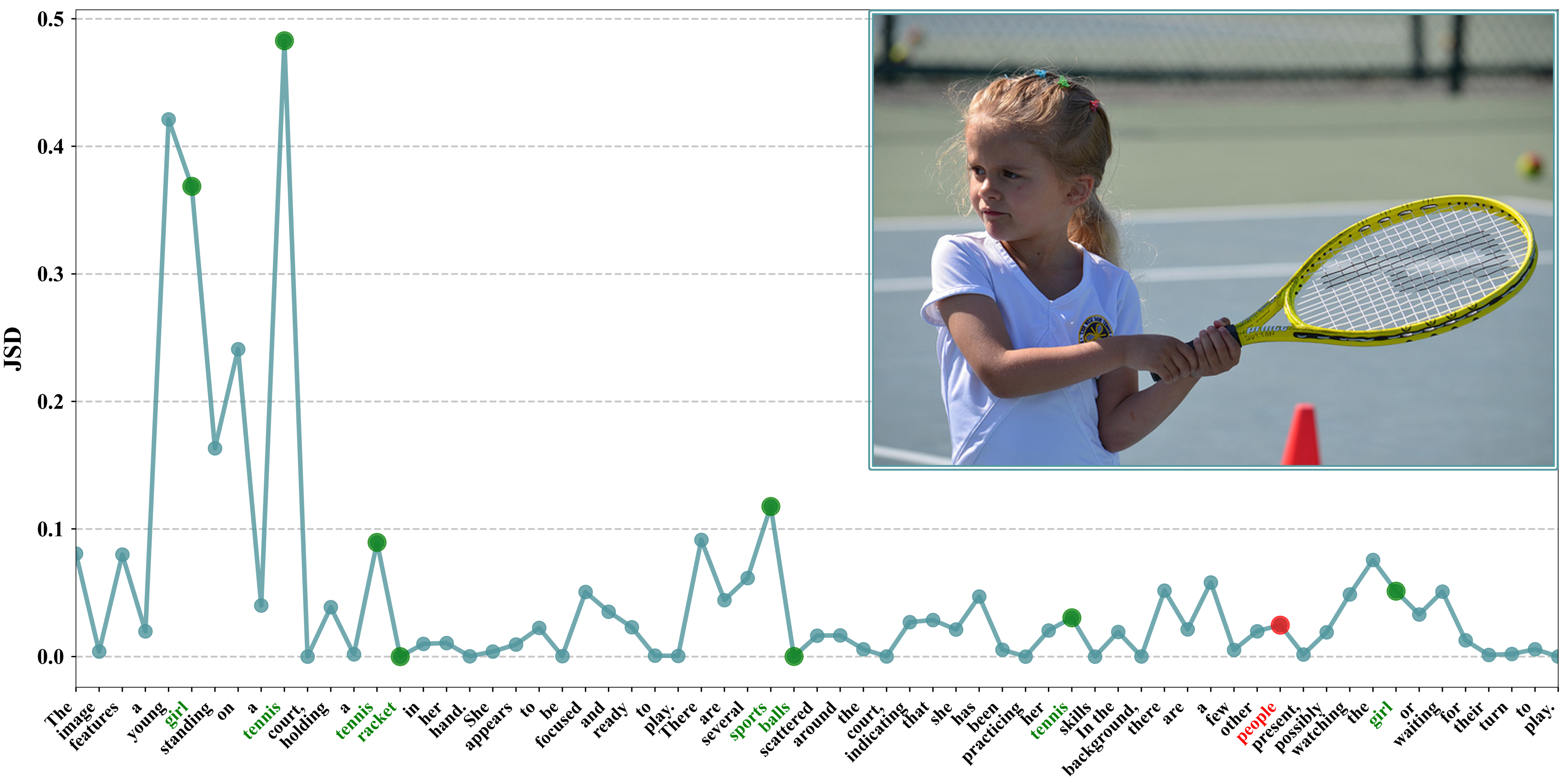} 
    \caption{
Illustration of JSD trend: hallucinated objects are marked in red and existent objects in green. Within the same position span, object words often exhibit higher JSD values than other words. Early-position object words yield high JSD values, whereas words in later positions consistently show low JSD values.
    } 
    \label{fig:jsd-case} 
\end{figure*}
\begin{figure}[t!] 
    \centering
    \includegraphics[width=0.90\linewidth]{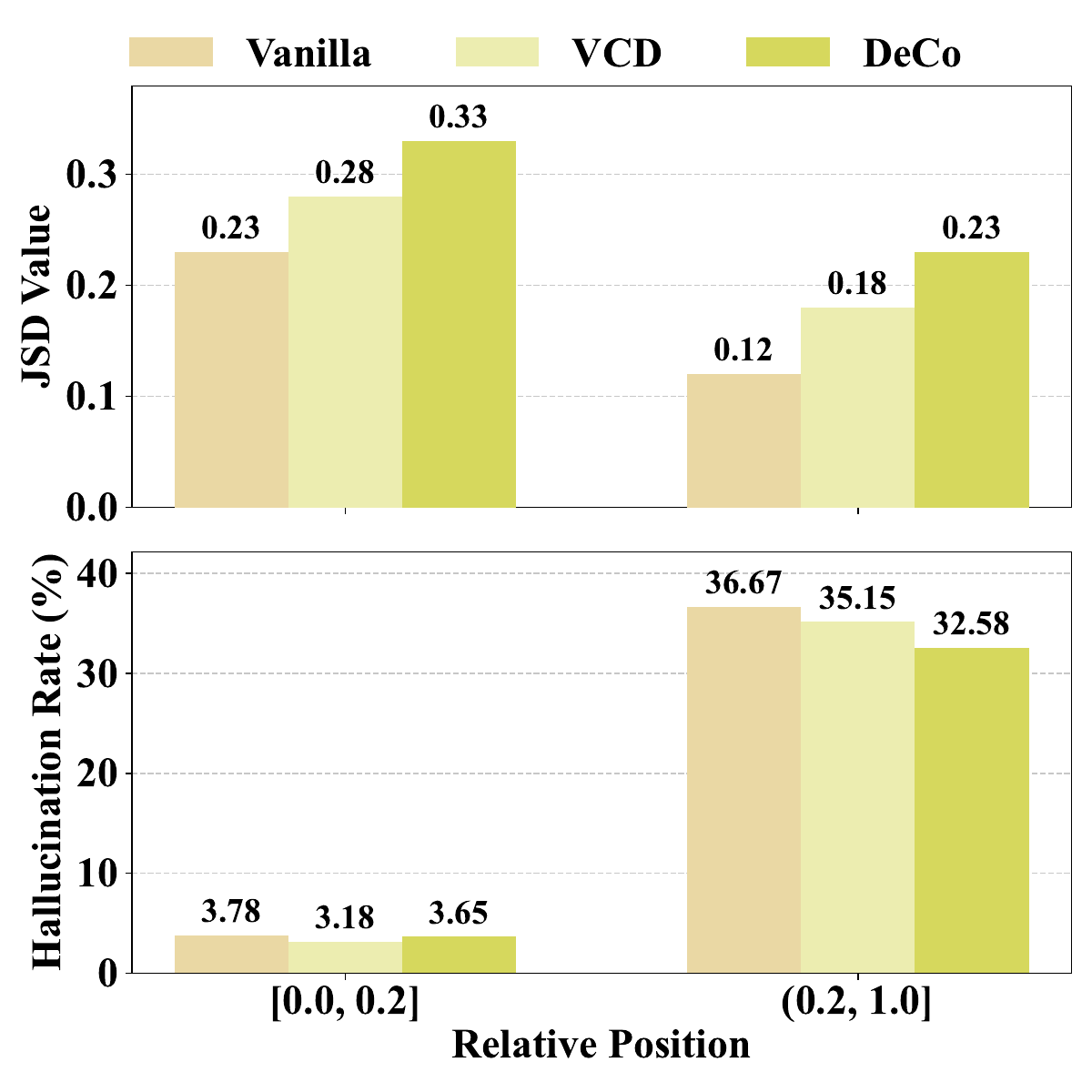} 
    \caption{ 
JSD value (top) and hallucination rate (bottom) for three methods across different relative position bins. As the generation progresses, JSD value decreases significantly, associated with a sharp increase in hallucination rate.
    } 
    \label{fig:jsd2} 
\end{figure}

\section{Introduction}
Recent advancements in Large Vision Language Models (LVLMs)~\cite{lmm-survey} have significantly enhanced their capacity to process and interpret visual inputs, achieving remarkable performance in tasks like image captioning and visual question answering~\cite{llava1.5, instructblip, Qwen2.5-VL}. However, these models remain prone to object hallucination, where the generated textual outputs diverge from the actual visual content, including objects not existing in image, undermining their reliability~\cite{chair, pope, nope}.
Addressing object hallucination is critical for real-world applications, as even minor inaccuracies can lead to misleading or nonsensical responses.

Extensive research has attributed the cause of object hallucination to LVLM's inherent \textbf{over-reliance on language priors}\textemdash where generation depends solely on previously generated textual context while ignoring the image~\cite{languagebias, vcd, vdd}. To mitigate this issue, recent work introduces logits calibration to suppress hallucinated logits or encourage factual ones. Among the most representative methods, VCD~\cite{vcd} contrasts output distributions derived from original and distorted visual inputs, while DeCo~\cite{deco} 
leverages preceding-layer to correct the final output distribution. 
However, our empirical analysis reveals that while these methods can mitigate the over-reliance to some extent, they fail to reverse the trend of increasingly severe over-reliance as generation length grows. 
Consequently, their object generation decisions still overly rely on language priors, resulting hallucinated object description.

Can we bypass this over-reliance trap to accurately estimate the confidence of object's existence? It is intuitive that answers consisting of only a single word or phrase can inherently avoid the problem, as they lack preceding generated text to rely on besides the image itself. Building on this insight, we propose a method termed \textbf{Language-Prior-Free Verification (LPFV)}. Following initial caption generation, LPFV is employed by instructing the model: ``\textit{Describe any element of the image with only one word or phrase.}''
The resulting distribution provides a confidence score for each object's existence.
Further experiments confirm this score is a more reliable verifier of object existence than the original object probabilities measured in caption. 
Consequently, LPFV serves as a powerful verifier, offering a direct mechanism to guide the improvement of factually grounded, less hallucinatory captions.


Building on these, we propose a novel \textbf{Self-Validation Framework} for mitigating object hallucination. It leverages LPFV to measure the existence confidence of objects in multiple candidate captions, offering two alternative strategies for producing final caption. The first strategy Best-of-N Selection selects one caption with the highest average confidence. The second strategy is Filter-then-Aggregate, where low-confidence object descriptions are first filtered out before the remaining reliable descriptions are aggregated to regenerate the final caption.

Extensive experiments demonstrate that self-validation framework significantly reduces object hallucination in LVLMs, surpassing state-of-the-art methods by a large margin while maintaining a favorable F1 score. Its robust performance under different hyperparameter configurations and strategy implementations validates its potential for practical applications. 

\section{Preliminary Study}

In this section, we first conduct a series of experiments to examine LVLMs' reliance on language priors when generating captions. 
Our analysis experiments are conducted in LLaVA-v1.5-7B~\cite{llava1.5}, based on samples randomly selected from the MSCOCO validation set~\cite{mscoco}.

\subsection{Preliminaries for LVLM Generation}
We consider an LVLM parameterized by $\theta$. The model typically takes a textual prompt $\mathbf{x}$ and an image $\mathbf{v}$ as inputs, and generates a response $\mathbf{y}$. The response is sampled  autoregressively from a probability distribution, where the generation decision at each step can be formulated as $ y_t  \sim p_\theta\left(y_t \mid {\mathbf{v}}, {\mathbf{x}}, {y}_{<t}\right).$

For each sampled caption $\mathbf{y}_i$, we can extract objects $\mathcal{O}_{\mathbf{y}_i} = \{\mathbf{o}_{1}, \ldots, \mathbf{o}_{j},\ldots, \mathbf{o}_{m_i}\}$, where $m_i$ is the total number of examined objects. 
Each object $\mathbf{o}_j$ is associated with its occurrence position in the caption, denoted by $s(\mathbf{o}_j)$.

\subsection{How Over-Reliance Leads to Hallucination}

To quantify the impact of language priors on LVLM's generation decision as the generation length increases, we conduct experiments to isolate the contribution of language priors to the generation decision.
Specifically, for each generation step, we measure \textbf{J}ensen-\textbf{S}hannon \textbf{D}ivergence (\textbf{JSD})~\cite{jsd} between two distributions as ${\text{JSD}\left(p_\theta (y_t \mid \mathbf{v}, \mathbf{x}, {y}_{<t})\parallel p_\theta (y_t \mid \mathbf{x}, {y}_{<t})\right)}$.
Since taking visual input $\mathbf{v}$ as a condition is the only difference between the two distributions,
a large discrepancy indicates a high contribution of visual input $\mathbf{v}$, while a minor discrepancy indicates the language priors $[\mathbf{x}, {y}_{<t}]$ dominate the generation.

To visualize the JSD trend in generating image caption, we select a sample and plot the JSD of each word generation in Figure~\ref{fig:jsd-case} (as a word can be decoded from multiple tokens, we assign the JSD value of its first token to the entire word).
The high JSD values of object words at early positions of a sentence confirm that the LVLM focuses on image when generating these words in the initial decoding steps.
However, after the first roughly 20\% of decoding steps, the JSD values drop sharply. This indicates a rapid accumulation of reliance on language priors, which quickly comes to dominate the generation process.



To further investigate the relationship between JSD and hallucination rate, we divide object words into two relative-position bins: an early bin [0, 0.2] and a late bin (0.2, 1.0]. This 0.2 threshold is chosen based on the observation that JSD typically drops sharply around this point. We randomly sample 100 instances and then calculate the average JSD and hallucination rate for each bin to analyze their correlation.
In addition to the vanilla greedy decoding method, our experiments include two recently proposed representative calibration methods for mitigating reliance on language-priors: VCD~\cite{vcd} and DeCo~\cite{deco}.

As illustrated in Figure~\ref{fig:jsd2}, the results reveal a clear inverse relationship between JSD values and hallucination rates. In the early bin [0.0, 0.2], all three methods exhibit high JSD values and correspondingly low hallucination rates.
However, in the late bin (0.2, 1.0], the JSD values for all methods drop significantly, while hallucination rates show a tenfold increase compared to the early bin. Although the calibration methods, VCD and DeCo, successfully maintain higher JSD values than the vanilla baseline, they fail to reverse the overall downward trend. This sharp decline in JSD, coupled with the surge in hallucination, underscores a critical insight: \textbf{over-reliance on language priors exacerbates with increasing generation length, making hallucinations increasingly likely to occur toward the later position of the output.}




\begin{figure}[t] 
    \centering
    \includegraphics[width=1.0\linewidth]{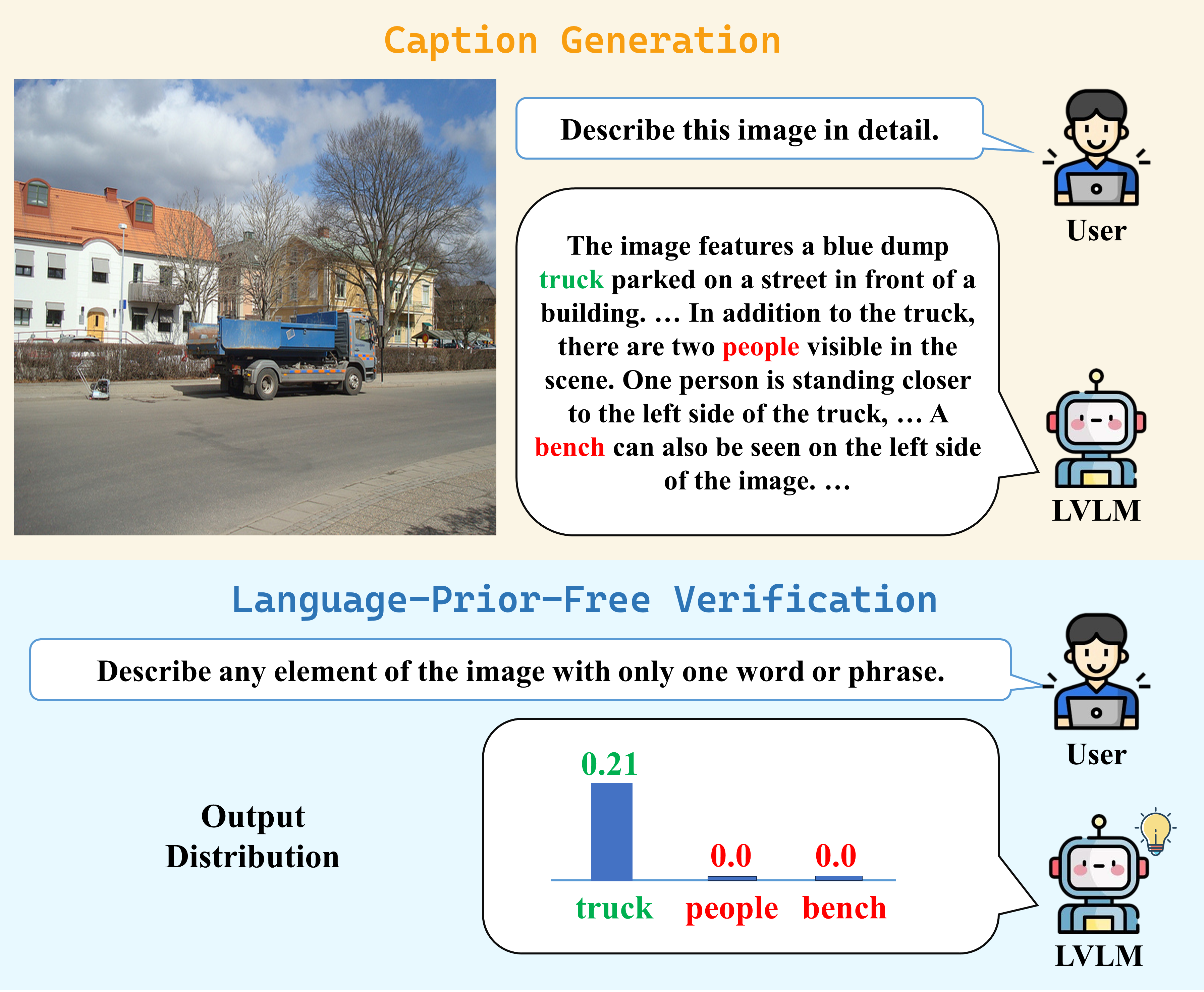} 
    \caption{
Hallucinated objects are highlighted in red while existent objects are in green. The LVLM provides a discriminative existence estimation of the objects when Language-Prior-Free Verification is employed, indicating that LVLMs have the ability to self-validate their generated objects.} 
    \label{fig:first} 
\end{figure}

\begin{figure*}[t]
\centering
\includegraphics[width=2.1\columnwidth]{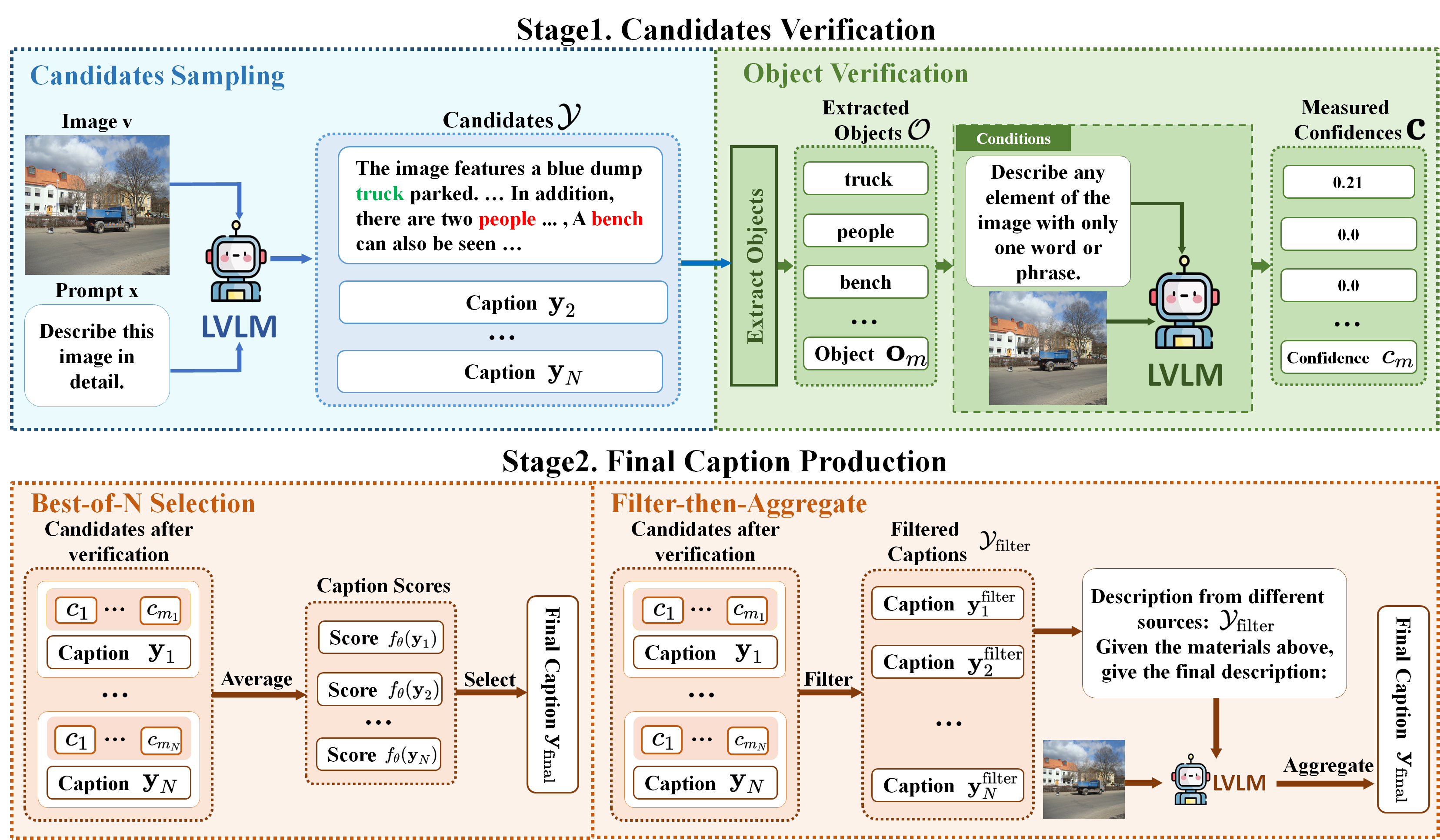}
\caption{
\textbf{Illustration of the Self-Validation Framework.} The framework operates in two stages. \textbf{Stage 1}: An LVLM first generates multiple candidate captions. For each candidate, the framework extracts objects and employs LPFV to assess their confidence scores. \textbf{Stage 2}: The final caption is produced via one of two strategies: (a) Best-of-N Selection, where the candidate with the highest confidence is chosen, or (b) Filter-then-Aggregate, where sentences with low-confidence objects are discarded before aggregating the remaining content.}

\label{fig:overview}
\end{figure*}

\section{Methods}
Inspired by our preceding finding, we first propose a novel verification method. Building upon this method, we then introduce a self-validation framework designed to mitigate object hallucination.

\subsection{Language-Prior-Free Verification}

Based on our finding that longer text generation exacerbates over-reliance on language priors, we propose \textbf{Language-Prior-Free Verification (LPFV)} to compel LVLMs to focus on the image rather than relying on previously generated text for verifying objects' existence. As illustrated in Figure~\ref{fig:first}, we prompt the LVLMs with instruction $\mathbf{x}_e$:

\vspace{0.5em}
\noindent\fbox{%
    \parbox{0.97\linewidth}{%
        \emph{Describe any element of the image with \textbf{only one word or phrase}}
    }%
}
\vspace{0.1em}

We expect LVLMs to output exactly one name of any existing object under the instruction $\mathbf{x}_e$.
Compared to the \textbf{original} object probability $p_\theta \left(\mathbf{o} \mid \mathbf{v}, \mathbf{x}, \mathbf{y}_{<s(\mathbf{o})}\right)$, \textbf{LPFV} measures object probability as $p_\theta \left(\mathbf{o} \mid \mathbf{v}, \mathbf{x}_e\right)$, eliminating the effect of language priors $\mathbf{y}_{<s(\mathbf{o})}$.

To compare the effectiveness of the two mentioned object probability (\textbf{original} and \textbf{LPFV}) in detecting hallucinated objects, we employ AUROC (Area Under the Receiver Operating Characteristic) for evaluation. It plots the true positive rate (TPR) against the false positive rate (FPR) at various thresholds and calculates the total area under this curve. A higher AUROC value indicates better discrimination performance, with 0.5 representing random guessing.
In our experiments, the proposed LPFV method achieves an AUROC of \textbf{0.85}, whereas the original probability yields an AUROC of \textbf{0.69}.\footnote{We also discuss leveraging the widely used CLIPScore for verification and plot the corresponding ROC curves in Appendix.} This comparison confirms that our LPFV delivers a more reliable estimation for distinguishing between hallucinated and non-hallucinated objects, revealing LVLMs' untapped potential when not being misled by language priors. The poor performance of the original object probability is corresponding to our finding that over-reliance on language priors leads to hallucination.


\begin{table*}[t]
\vspace{-4pt}
\centering
\resizebox{1.0\textwidth}{!}{
\begin{tabular}{@{}c|l|l|c|ccc|cc}
\toprule
Row & Model & Method & Length & CHAIR\(_S\) $\downarrow$ & CHAIR\(_I\) $\downarrow$ & F1 $\uparrow$ & Acc. $\uparrow$ & Rel. $\uparrow$ \\ \midrule
1 & \multirow{10}{*}{LLaVA-v1.5-7B} & - & 100.6 & 50.0 & 15.4 & 75.1 & 7.50 & 7.89 \\
2 &  & VCD \scriptsize{[CVPR24]} & 100.4 & 48.6 & 14.9 & 75.2 & 7.54 & 8.13 \\
3 &  & CGD \scriptsize{[ICLR24 workshop]} & 89.6 & 37.2 & 11.4 & \textbf{78.2} & 7.76 & 8.23 \\
4 &  & HALC \scriptsize{[ICML24]} & 72.4 & 32.2 & 10.9 & 69.0 & 6.37 & 6.51 \\
5 &  & Less \scriptsize{[ACL24]} & 79.0 & 38.8 & 12.0 & 77.6 & 7.64 & 8.07 \\
6 &  & DeCo \scriptsize{[ICLR25]} & 99.0 & 42.4 & 12.9 & 73.3 & 6.96 & 7.66 \\
7 &  & Nullu \scriptsize{[CVPR25]} & 99.9 & 49.4 & 13.5 & 76.6 & 7.71 & 8.18 \\
8 &  & \textbf{Self-Val.~\scriptsize{(w/ BoN.$N$=10)}} & 98.2 & 28.8 & 7.3 & 76.7 & 7.75 & 8.26\\
9 &  & \textbf{Self-Val.~\scriptsize{(w/ FtA.$N$=3)}} & 115.2 & \textbf{22.8} & \textbf{5.3} & 76.1 & \textbf{7.80} & \textbf{8.30} \\
\midrule
10 & \multirow{3}{*}{LLaVA-1.5-13B} & - & 101.0 & 45.8 & 13.0 & 77.0 & 7.65 & 7.29 \\
11 &  & \textbf{Self-Val.~\scriptsize{(w/ BoN.$N$=10)}} & 97.2 & 24.0 & 6.6 & \textbf{77.2} & 7.72 & 8.20 \\
12 &  & \textbf{Self-Val.~\scriptsize{(w/ FtA.$N$=3)}} & 121.7 & \textbf{21.2} & \textbf{5.1} & 76.7 & \textbf{7.96} & \textbf{8.32} \\ \midrule

13 & \multirow{3}{*}{mPLUG-Owl2-7B} & - & 105.8 & 60.8 & 18.9 & 71.6 & 7.28 & 7.66 \\
14 &  & \textbf{Self-Val.~\scriptsize{(w/ BoN.$N$=10)}} & 101.2 & 34.2 & 9.4 & \textbf{74.5} & \textbf{7.38} & \textbf{7.96} \\
15 &  & \textbf{Self-Val.~\scriptsize{(w/ FtA.$N$=3)}} & 98.4 & \textbf{13.2} & \textbf{4.2} & 70.5 & 7.22 & 7.76 \\ \midrule
16 & \multirow{3}{*}{Qwen2.5-VL-7B} & - & 175.3 & 35.6 & 9.9 & \textbf{74.4} & 8.52 & 8.70 \\
17 &  & \textbf{Self-Val.~\scriptsize{(w/ BoN.$N$=10)}} & 172.8 & 16.6 & 4.1 & 72.7 & \textbf{8.94} & \textbf{9.13} \\
18 &  & \textbf{Self-Val.~\scriptsize{(w/ FtA.$N$=3)}} & 156.9 & \textbf{8.4} & \textbf{3.1} & 70.6 & 8.57 & 8.81 \\ 

\bottomrule
\end{tabular}
}

\caption{\label{tab:main-res}
\textbf{Object hallucination performance across different methods and models on image captioning task.} Best results are in \textbf{bold}. \textbf{w/ BoN.} and \textbf{w/ FtA.} denote Best-of-N Selection and Filter-then-Aggregate. \textbf{Acc.} and \textbf{Rel.} stand for the GPT-assisted metrics Accuracy and Relevancy.
}

\end{table*}

\subsection{Self-Validation Framework}
Based on LPFV, we introduce a self-validation framework to mitigate object hallucination in generated captions. As illustrated in Figure~\ref{fig:overview}, the framework operates in two stages. In \textbf{Stage 1: Candidate Verification}, it leverages LPFV to identify potential hallucinations among multiple candidate captions. Subsequently, in \textbf{Stage 2: Final Caption Production}, it produces more factually grounded caption using either Best-of-N Selection or Filter-then-Aggregate strategy.

\label{sec:candiates}
\subsubsection{3.2.1 Candidates Verification}
\paragraph{Candidates Sampling.}
We follow the standard approach of sampling $N$ captions via multinomial sampling. 
Each sampling process is defined as $\mathbf{y} \sim p_\theta\left(\mathbf{y} \mid \mathbf{v},\mathbf{x}\right)$,
where $\mathbf{v}$ is the given image and $\mathbf{x}$ is the instruction of \textit{``Please Describe this image in detail.''} Then, we collect $N$ candidate captions as $\mathcal{Y} = \{\mathbf{y}_1, ...,\mathbf{y}_i, ..., \mathbf{y}_N\}$.

\paragraph{Object Verification.}
For each candidate caption $\mathbf{y}_i$, we first extract its object words $\mathcal{O}_{\mathbf{y}_i} = \{\mathbf{o}_{1},..., \mathbf{o}_{j},..., \mathbf{o}_{m_i}\}$. For every $\mathbf{o}_j \in \mathcal{O}_{\mathbf{y}_i}$, LPFV is employed to get corresponding confidence score $c_j = p_\theta\left(\mathbf{o}_j \mid \mathbf{v}, \mathbf{x}_e \right)$.
The object confidences of the candidate is $\mathbf{c}_{_i}=\{{c}_1,...,{c}_j,...,{c}_{m_i}\}$.

\subsubsection{3.2.2 Final Caption Production}
\label{sec:final}

\paragraph{Best-of-N Selection.}
After the above steps, a direct option is to select one from the sampled candidates $\mathcal{Y}$ as the final response $\mathbf{y}_\text{final}$. 
For each candidate $\mathbf{y}_i$, we average the confidence of its extracted objects as the caption-level score, defined as:
\begingroup
\thickmuskip=1.4mu  
\begin{equation}
    \begin{aligned}
    f_{\theta}(\mathbf{y}_i) = \frac{1}{m_i}\sum_{j=1}^{m_i}c_{j}
    =\frac{1}{m_i}\sum_{j=1}^{m_i}p_{\theta}\left(\mathbf{o}_j \mid \mathbf{v}, \mathbf{x}_e\right),
    \end{aligned}
\end{equation}
\endgroup
where $c_j \in \mathbf{c}_i$ and $\mathbf{o}_j \in \mathcal{O}_{\mathbf{y}_i}$.
Then we select the one with the highest caption-level score as the final caption.

However, we figure out that selection does not fully exploit the potential of our framework due to the following reasons:
(1) The selected caption can still contain hallucinated objects.
(2) The complementary factual descriptions among different candidates have not been fully utilized.

\paragraph{Filter-then-Aggregate.}
To minimize hallucinated content and fully utilize the complementary descriptions across candidates, we propose a Filter-then-Aggregate strategy for regenerating final captions. 

In the filtering stage, we discard sentences containing potentially hallucinated objects (confidence score $c_j \leq \alpha$) for each candidate $\mathbf{y}_i$, obtaining filtered caption $\mathbf{y}^{\text{filter}}_{i}$. The resulting set of filtered candidates is denoted as $\mathcal{Y}_\text{filter}$.

Then, we prompts the LVLM to aggregate the filtered candidates to regenerate a final caption $\mathbf{y}_\text{final}$. The prompt for aggregation is defined as:

\vspace{0.3em}
\noindent\fbox{%
    \parbox{0.97\linewidth}{%
        \emph{Description from different sources: $\mathcal{Y}_\text{filter}$}\\
\emph{Given the materials above, 
give the final description: }
    }%
}
\vspace{0.1em}

The filtering stage enables the faithfulness of the candidate captions, and the aggregation stage can fully exploit the facts in candidates for regenerating the final caption.

\section{Experiments}
\subsection{Setup}
\noindent \textbf{Models.} As our self-validation framework can be seamlessly applied to any LVLMs, we choose LLaVA-v1.5~\cite{llava1.5}, mPLUG-Owl2~\cite{ye2023mplug} and Qwen2.5-VL~\cite{Qwen2.5-VL} as the base models.

\noindent \textbf{Baselines.}
To demonstrate the effectiveness of our framework, we evaluate it against SOTA decoding methods including VCD~\cite{vcd}, CGD~\cite{cdg}, HALC~\cite{halc}, DeCo~\cite{deco}, as well as the training-based method Less~\cite{less} and the editing-based method Nullu~\cite{nullu}. All of these methods are designed for object hallucination mitigation.
To ensure a fair comparison, we standardize the decoding configurations across all methods. Specifically, we set the maximum number of new tokens to 512 to guarantee complete caption generation and employ Top-k sampling with k=50 to avoid selecting low-probability tokens. The temperature is set to 0 for generation, except during the candidate sampling stage, where we apply a temperature of 0.5 to enhance diversity. We reproduce these methods to evaluate their performance; full implementation details are provided in Appendix.

\noindent \textbf{Evaluation.} 
Following previous work, we randomly sample 500 images from the MSCOCO validation set to conduct experiments~\cite{mscoco,less}. 
Specifically, we query different LVLM models with the same prompt \textit{``Please describe this image in detail.''} 

For evaluation metric, we follow previous works~\cite{hallucidoctor} to use Caption Hallucination Assessment with Image Relevance (CHAIR)~\cite{chair}. 
It quantifies the hallucination at both the instance level (CHAIR\textsubscript{I}) and sentence level (CHAIR\textsubscript{S}). These two metrics are defined as the average results of:
$$
\text{CHAIR}_I = \frac{\#\text{hallucinated objects}}{\#\text{all mentioned objects}},
$$
$$
\text{CHAIR}_S = \frac{\#\text{captions with hallucinated objects}}{\#\text{all captions}}
.
$$

As the CHAIR metrics can be hacked by suppressing the richness and completeness of the captions (e.g., generating shorter or meaningless answers), 
we follow previous work to incorporate F1 score for evaluation~\cite{less, pai}, which is defined as:
$$
\text{F1} = \frac{2 * \text{precision} * \text{recall}}{\text{precision} + \text{recall}}
.
$$
Since CHAIR and F1 only extract objects in captions for rule-based evaluation, we also incorporate GPT-assisted evaluation to assess overall caption quality. Specifically, we leverage GPT-4o-mini to perform a two-faceted evaluation. Each caption is scored from 1 to 10 on Accuracy (Acc.), which quantifies its factual correctness by penalizing content inconsistent with the image, and Relevancy (Rel.), which measures the coverage of the image's core elements and main subjects. Detailed prompt templates and scoring criteria can be found in Appendix.


\subsection{Experimental Results}
Experimental results in Table~\ref{tab:main-res} demonstrate ours effectiveness in reducing hallucinations across LVLMs. When applied to LLaVA-v1.5-7B~\cite{llava1.5}, it achieves significant hallucination reduction (52.2\%--65.6\% decrease in $\text{CHAIR}_I$) across both strategies, without sacrificing the richness of the descriptions. 
Compared to other hallucination-mitigation methods on LLaVA-v1.5-7B, our method sets new state-of-the-art results for object hallucination. Moreover, it achieves consistent and substantial reductions in object hallucination across four LVLMs and two model sizes (7B and 13B).
Regarding GPT-assisted evaluation metrics, our framework consistently outperforms the baselines. 

Within our framework, Filter-then-Aggregate (FtA) always achieves a lower hallucination rate than Best-of-N (BoN) with a slight decrease in F1. This can be attributed to its mechanism of discarding descriptions of potentially hallucinated objects. However, BoN still holds an advantage on GPT-assisted evaluation in two of the four models. We hypothesize this stems from its adherence to the original distribution, thereby better preserving faithfulness. To enable a more thorough comparison between these two strategies, we provide qualitative results in Appendix.


\subsection{Analysis Study}
\begin{figure}[t]
    \centering
    \begin{subfigure}[b]{0.49\linewidth}
        \centering
        \includegraphics[width=\linewidth]{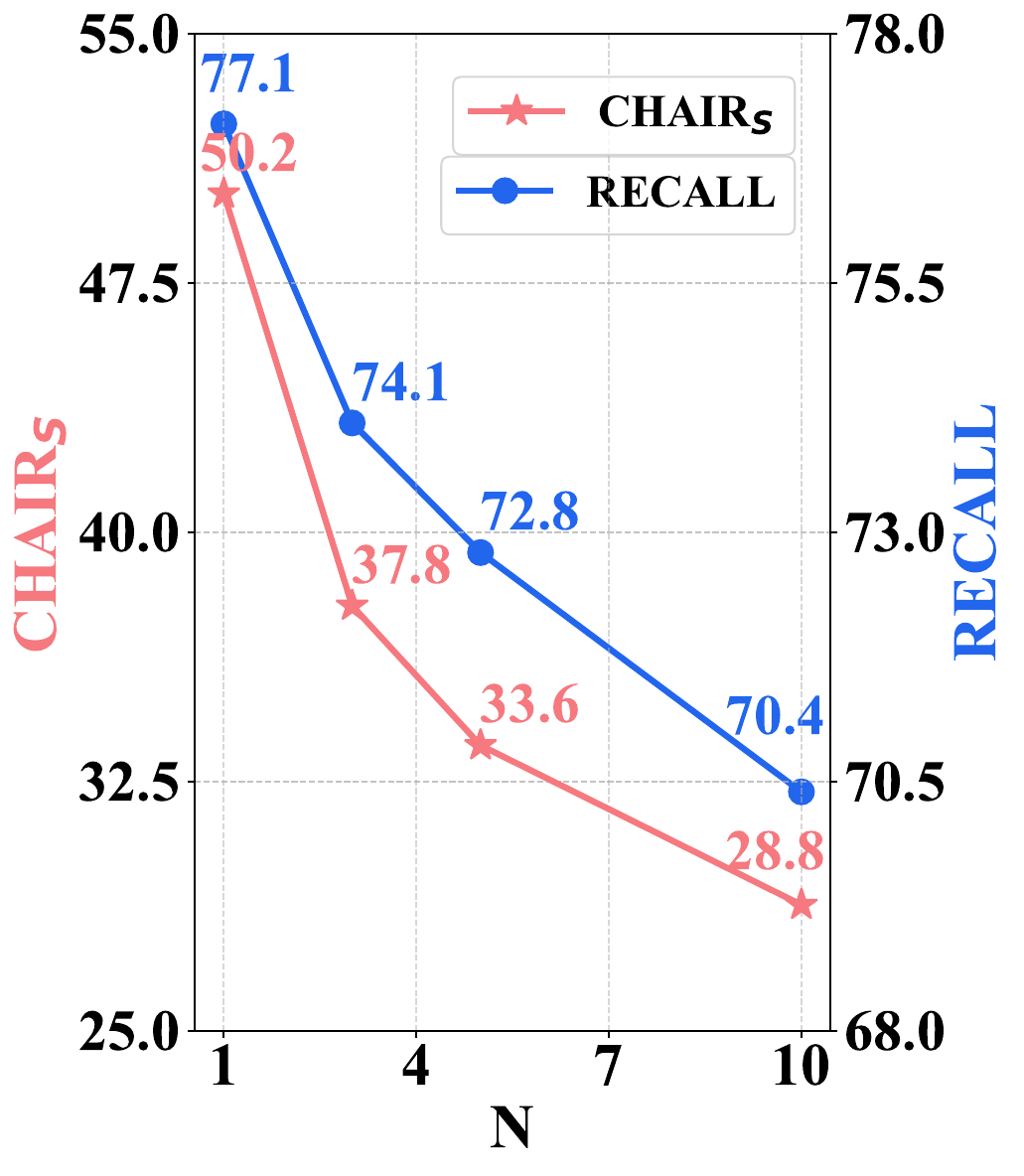} 
        \caption{Best-of-N Selection}
        \label{fig:BoN}
    \end{subfigure}
    \hfill
    \begin{subfigure}[b]{0.49\linewidth}
        \centering
        \includegraphics[width=\linewidth]{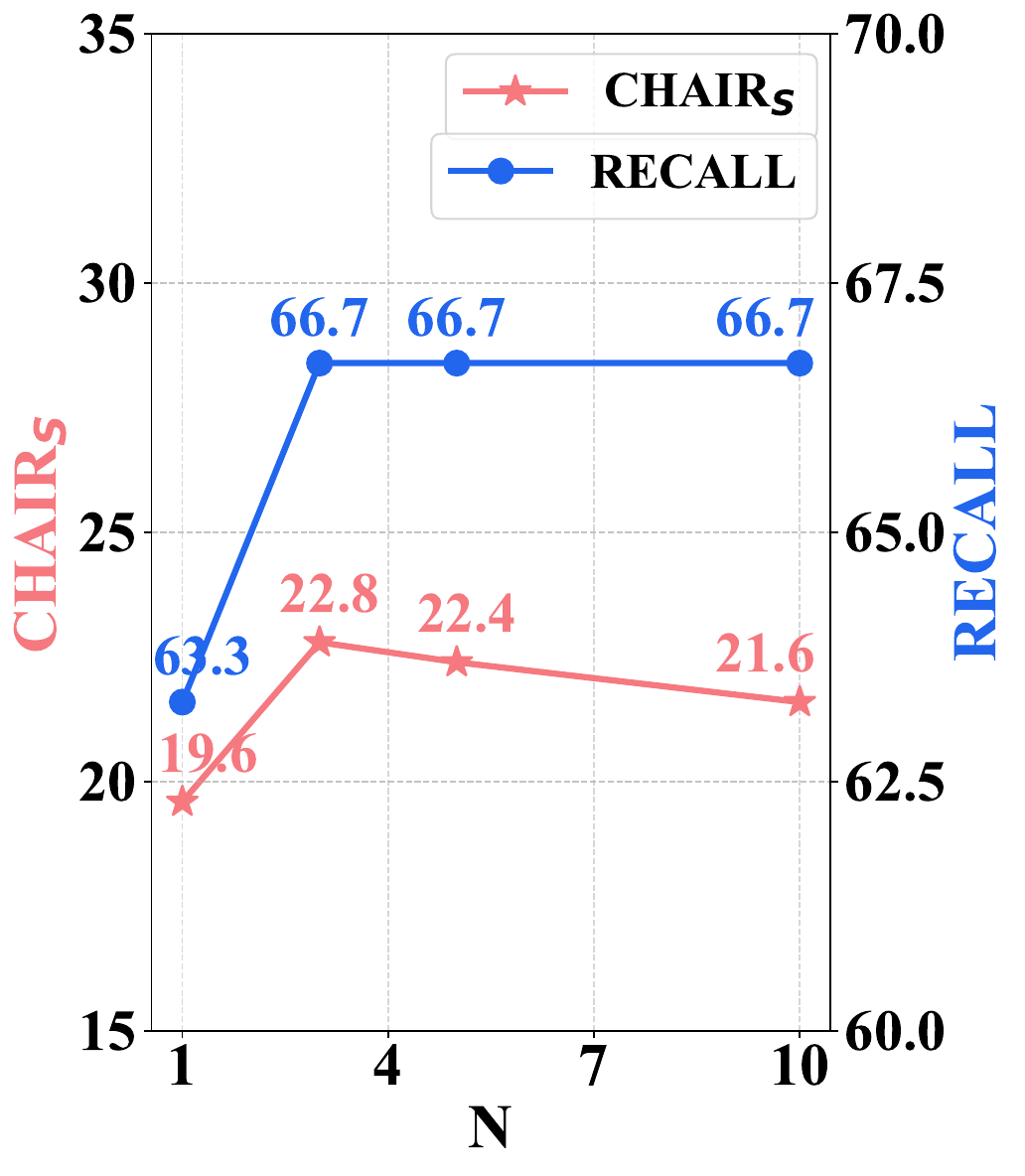} 
        \caption{Filter-then-Aggregate}
        \label{fig:FtA}
    \end{subfigure}
    
    \caption{Impact of different candidates nums $N$ on hallucination and recall performance.}
    \label{fig:candidates}
    
\end{figure}

\noindent \textbf{Effect of Candidates Number.}
Our framework involves a key parameter $N$, which determines the number of generated caption candidates. For the Best-of-N strategy, a larger $N$ expands the candidate pool, increasing the likelihood of selecting captions with less hallucination. However, this may also lead to potentially overlooking captions with more rich description. For the Filter-then-Aggregate strategy, a larger candidate set provides richer descriptions for aggregation, while simultaneously introducing the risk of incorporating hallucinated content. Therefore, to thoroughly analyze the effect of $N$, we measure performance using both $\text{CHAIR}_I$ and recall.
We evaluate both strategies with $N \in \{1,3,5,10\}$ on LLaVA-v1.5-7B. For Filter-then-Aggregate, we fix the filter threshold $\alpha = 0.01$. Results are shown in Figure~\ref{fig:candidates}. Our key findings are as follows:
\begin{itemize}
    \item For BoN, increasing $N$ leads to a substantial reduction in the hallucination rate (from 50.2 to 28.8), while at the cost of a moderate decrease in recall (from 77.1 to 70.4).
    \item For FtA, although a larger $N$ increases the number of description details to be aggregated (potentially raising hallucination risk), the filtering mechanism effectively prevents significant hallucination growth. Meanwhile, more candidates facilitate the aggregation of more details, thereby improving recall from 63.3 to 66.7. Notably, the performance stabilizes when $N \geq 3$.
\end{itemize}

\begin{figure}[htbp] 
    \centering
    \includegraphics[width=\linewidth]{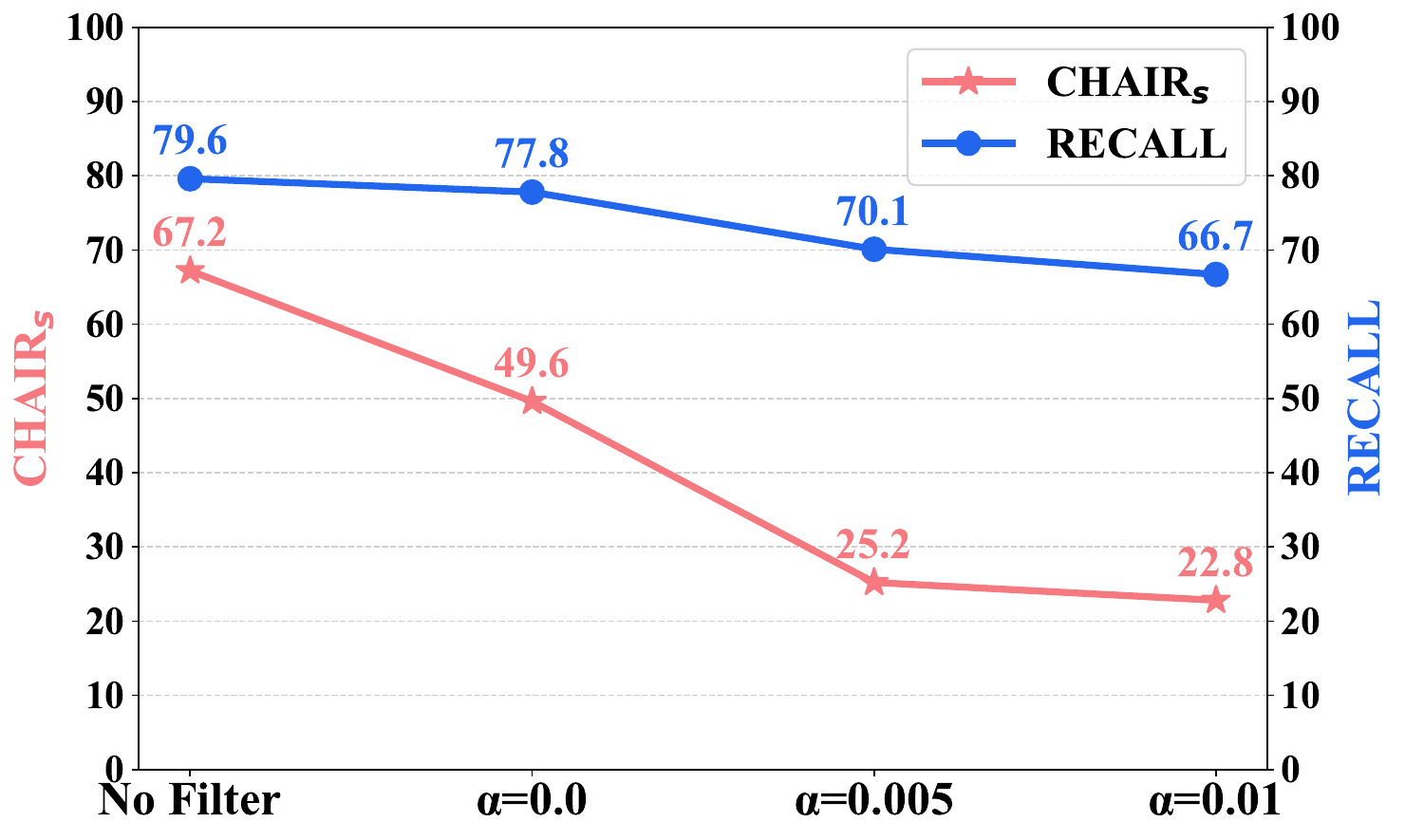} 
    \caption{Impact of different filter threshold $\alpha$ on hallucination and recall performance.}
    \label{fig:filter-ratio}
\end{figure}

\noindent \textbf{Effect of Filter Threshold.}
The filter threshold $\alpha$ is another key hyperparameter in our framework. Specifically, we discard any sentence containing object $\mathbf{o}_j$, if its confidence score $c_j \leq \alpha$. We analyze the framework's sensitivity to $\alpha$ across three values: $\alpha \in \{0, 0.005, 0.01\}$. Besides, we also compare these settings against a no-filter baseline that directly aggregates candidates to validate the effectiveness of our filtering mechanism.
It is important to note that our Top-k (k=50) sampling process inherently assigns $c_j=0$ to objects outside the top 50, meaning that $\alpha=0$ setting already represents a basic level of filtering.
All experiments are conducted with a fixed candidate number $N=3$ (results are shown in Figure~\ref{fig:filter-ratio}).
Our key findings are summarized below: 
\begin{itemize}
    \item Direct aggregation without filtering exacerbates hallucination by incorporating more potentially hallucinated descriptions, highlighting the critical role of filtering.
    \item A higher $\alpha$ filters out more likely-hallucinated descriptions. This yields a substantial reduction in $\text{CHAIR}_I$ from 49.6 to 22.8, which is far more significant than the decrease in recall from 77.8 to 66.7, showing the benefit of increasing $\alpha$.
\end{itemize}



\begin{figure}[t] 
    \centering
    \includegraphics[width=1.0\linewidth]{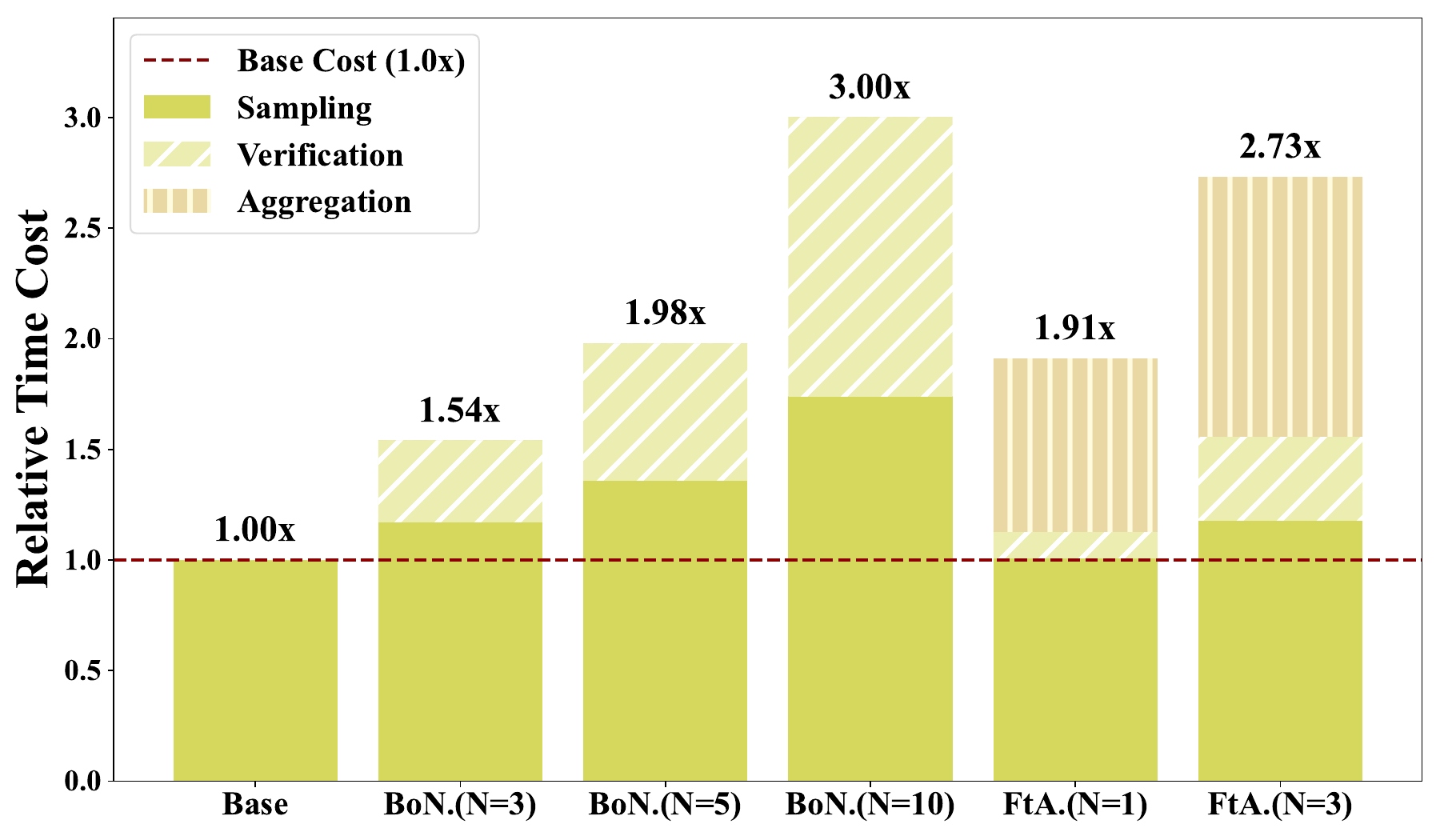} 
    \caption{Time cost analysis of our framework with different values of $N$ and strategies when applied to LLaVA-v1.5-7B.}
    \label{fig:speed}
\end{figure}

\noindent \textbf{Time Cost Analysis.}
Compared with vanilla greedy decoding, our framework incurs additional inference overhead. The extra latency stems from three sources: (1) \textbf{candidates sampling}, which is a parallelizable process; (2) \textbf{verification}, which requires a limited number of additional inference steps; and (3) \textbf{aggregation phase} unique to the FtA strategy. The time cost analysis presented in Figure~\ref{fig:speed} shows that, latency rises with the number of candidates $N$, and for the same $N$, BoN is faster than FtA.



\noindent \textbf{Comparison of Object Extraction Methods.}
The object extraction step is crucial for our framework as it directly impacts two components: (1) candidates scoring in Best-of-N Selection and (2) sentence filtering decision in Filter-then-Aggregate. 
Our default implementation employs a rule-based matching approach based on a manually predefined object set derived from \citet{chair}. This set comprises 80 common objects along with their various synonyms, yielding approximately 500 object nouns in total.

To enhance the generalization of our method for scenarios without predefined object sets,
we also explore to leverage the LVLM itself for object extraction. Based on whether extraction occurs during testing, we define two modes: \textit{offline} and \textit{online}. For \textit{offline} mode, we first randomly sample 2,000 captions from MSCOCO training set~\cite{mscoco}, and then prompt the LVLMs with few-shot examples to extract objects from each caption (see Appendix for prompt details). We then select the top 1,000 most common object nouns, which are used for rule-based matching during testing. For \textit{online} mode, the LVLMs are prompted during testing to extract objects from captions.

Results of our framework under the two modes in LLaVA-v1.5-7B are presented in Table~\ref{tab:extract-results}. The hyper-parameter setting is consistent to main experiments.
Although these modes underperform the predefined object set, they remain effective at reducing hallucinations, particularly achieving comparable performance when applied to FtA.
Due to the small LVLM's limited instruction-following capability, the online extraction may lead to imprecise object extraction. This results in a less pronounced reduction in the hallucination rate compared to the offline mode. However, it gains a comparable or even better performance in GPT-assisted evaluation for its more flexible and general object extraction.
Additionally, since it requires no prior in-domain data preparation, this approach offers broader applicability.

\begin{table}[t]
  \centering
    \begin{tabular}{ccccccc}
    \toprule
    Strategy & Mode & $C_S$ & $C_I$ & F1 & Acc. & Rel. \\
    \midrule
    \multirow{2}{*}{BoN.} 
      & \textit{Online} & 45.0 & 12.1 & 75.7 & \textbf{7.60} & 8.06 \\
      & \textit{Offline} & \textbf{41.8} & \textbf{10.9} & \textbf{75.9} & \textbf{7.60} & \textbf{8.11}\\
    \midrule
    \multirow{2}{*}{FtA.}
      & \textit{Online} & 30.4 & 7.6 & \textbf{76.5} & \textbf{7.45} & \textbf{8.02}\\
      & \textit{Offline} & \textbf{24.0} & \textbf{6.6} & 75.0 & 7.16 & 7.65\\
    \bottomrule
    \end{tabular}
  \caption{Performance comparison of online and offline extraction modes across the two caption production strategies on LLaVA-v1.5-7B.}
  \label{tab:extract-results}
\end{table}%

\section{Related Work}
\subsection{Causes of LVLM Hallucination}
Recent studies have identified multiple factors contributing to hallucination in LVLMs.
A key bottleneck lies in the visual encoder's inability to capture fine-grained details, restricting LVLMs' capacity for accurate visual details understanding~\cite{comm, mmvp, haelm}. 
Another source of hallucination arises from the inherent bias in LVLMs. Models tend to generate high-confidence responses based on language priors or statistical bias rather than visual evidences~\cite{lure, vcd, pai, infact}. Additionally, \citet{less} find that overly detailed training data hinders LVLMs' end-of-sequence (EOS) decision ability, resulting in hallucinated descriptions.


\subsection{Mitigating Hallucination for LVLMs}
Recent work on mitigating hallucinations in LVLMs can be categorized into three kinds: training-phase strategies, decoding-stage strategies, and post-hoc correction.
Training-phase strategies primarily focus on improving model optimization objectives or constructing more robust training datasets. Notably, \citet{ha-dpo} and \citet {chip-dpo} leverage specialized DPO variants, while \citet{less} propose selective EOS supervision for training. \citet{lrv-instruction} construct robust visual instruction dataset for mitigating hallucination.  
Current decoding strategies for hallucination mitigation primarily focus on suppressing hallucinatory logits or enhancing truthful logits through calibration. As a prominent approach, contrastive decoding deliberately transforms inputs to induce hallucinations and then contrast them~\cite{vcd, vdd, sid}, while \citet{deco} utilize the logits from lower-layer to calibrate the final logits. 

Other works adopt post-hoc correction and decoding-time interventions to rectify hallucinated content in model outputs. For instance, \citet{lure} utilizes a trained revisor, while \citet{woodpecker} integrates external detection models to refine the generated captions. Beyond relying on external modules, recent studies have delved into the internal dynamics of LVLMs for hallucination mitigation. For example, VDC \cite{VDC} analyzes the generation dynamics and rectifies errors by detecting unsupported subdominant tokens and replacing them with validated dominant ones.

In parallel, attention-based methods have been proposed to enhance visual grounding without the significant computational overhead of contrastive decoding or auxiliary models~\cite{pai,PADE}. 

However, despite the success of these external revisors, token-level replacements, and attention interventions, they largely overlook the fundamental bias introduced by textual context. Compared to existing works, our framework is the first to unlock the inherent capability of LVLMs to self-detect and rectify hallucinations by explicitly overcoming the misleading effects of language priors.

Compared to existing work, our framework is the first to unlock LVLMs' ability to detect and improve hallucination through overcoming the over-reliance on language priors. 

\section{Conclusions}
In this paper, we conduct a systematic analysis of the fundamental cause of hallucinations in LVLMs\textemdash over-reliance on language priors, finding that this problem becomes more severe as generation length increases. To counter this over-reliance trap, we propose a novel strategy to accurately estimate object confidence scores. We further integrate it into our self-validation framework to produce final captions with less hallucination. Experiments demonstrate that our framework achieves state-of-the-art performance in object hallucination mitigation, providing new practical directions for advancing this research field.

\bibliography{aaai2026}

\appendix

\section{Different Verification Methods}
\subsection{ROC Curves}

In addition to the original object probability and LPFV, CLIPScore could also serve as an option for measuring objects' existence confidence. It avoids over-reliance on language priors by directly computing the cosine similarity between visual and textual representations. However, as a sentence-level metric, CLIPScore represents the image in a relatively coarse way\textemdash different objects within the same sentence are forced to share the same score. The scoring of each object can be disproportionately influenced by other object descriptions in the same sentence.
Therefore, we also compute object-level CLIPscore by matching object names with the image. 

\begin{figure}[h!] 
    \centering
    \includegraphics[width=\linewidth]{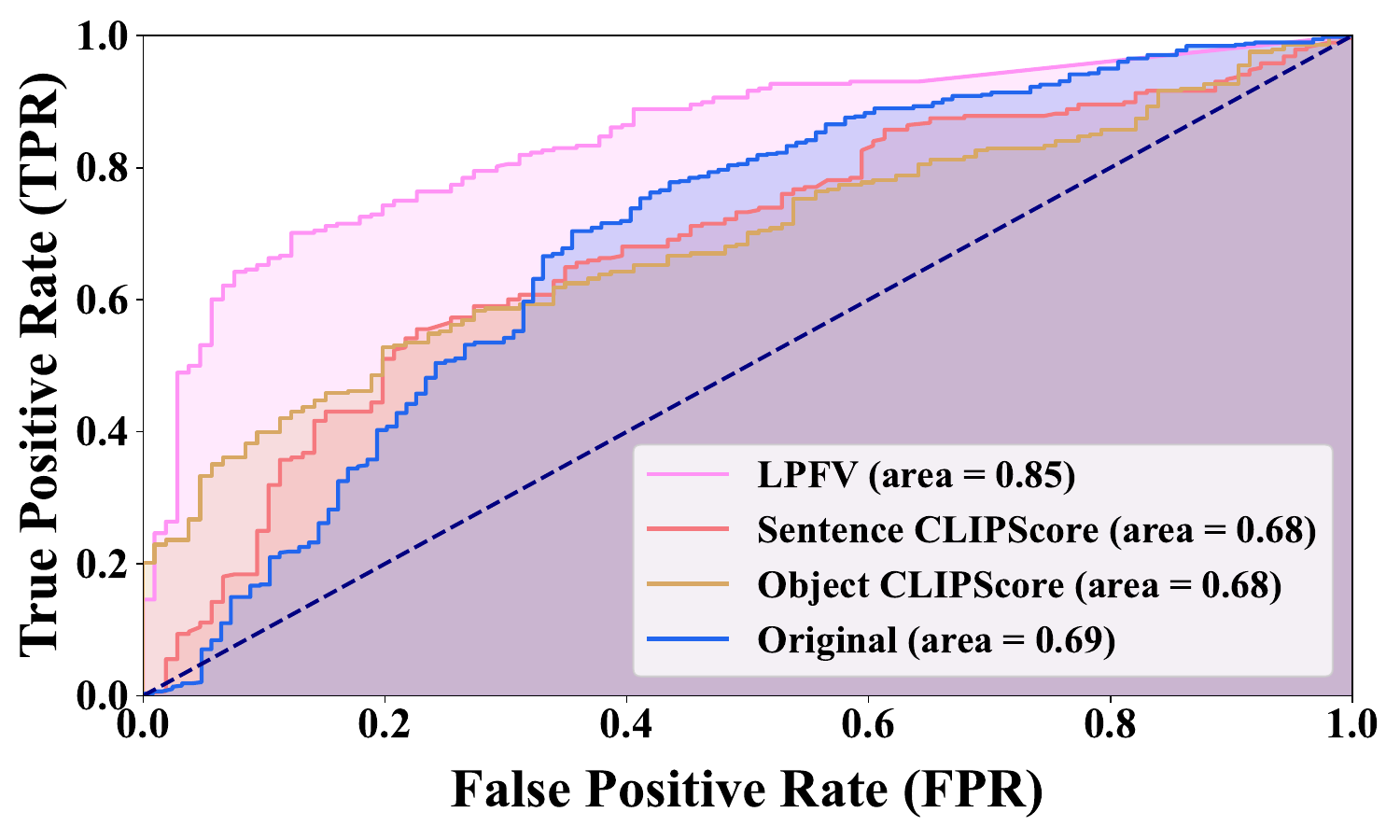} 
    \caption{ROC curves of different verification methods on LLaVA-v1.5-7B. Higher AUROC indicates better performance, with 0.5 representing random guessing and 1.0 being perfect discrimination.}
    \label{fig:ROC_curve}
    \vspace{-0.3em}
\end{figure}

As illustrated in Figure~\ref{fig:ROC_curve}, our proposed LPFV strategy most accurately distinguishes between hallucinated and non-hallucinated objects, revealing the untapped potential of LVLMs once freed from misleading language priors. The poor performance of using the original object probability reinforces our finding that language priors dominate generations. Furthermore, we attribute the limited effectiveness of CLIPScore to the fundamental challenge of handling multiple objects within one image.

\subsection{CLIPScore for Object Verification}
\label{app:tradeoff}
\begin{figure}[h!] 
    \centering
    \includegraphics[width=\linewidth]{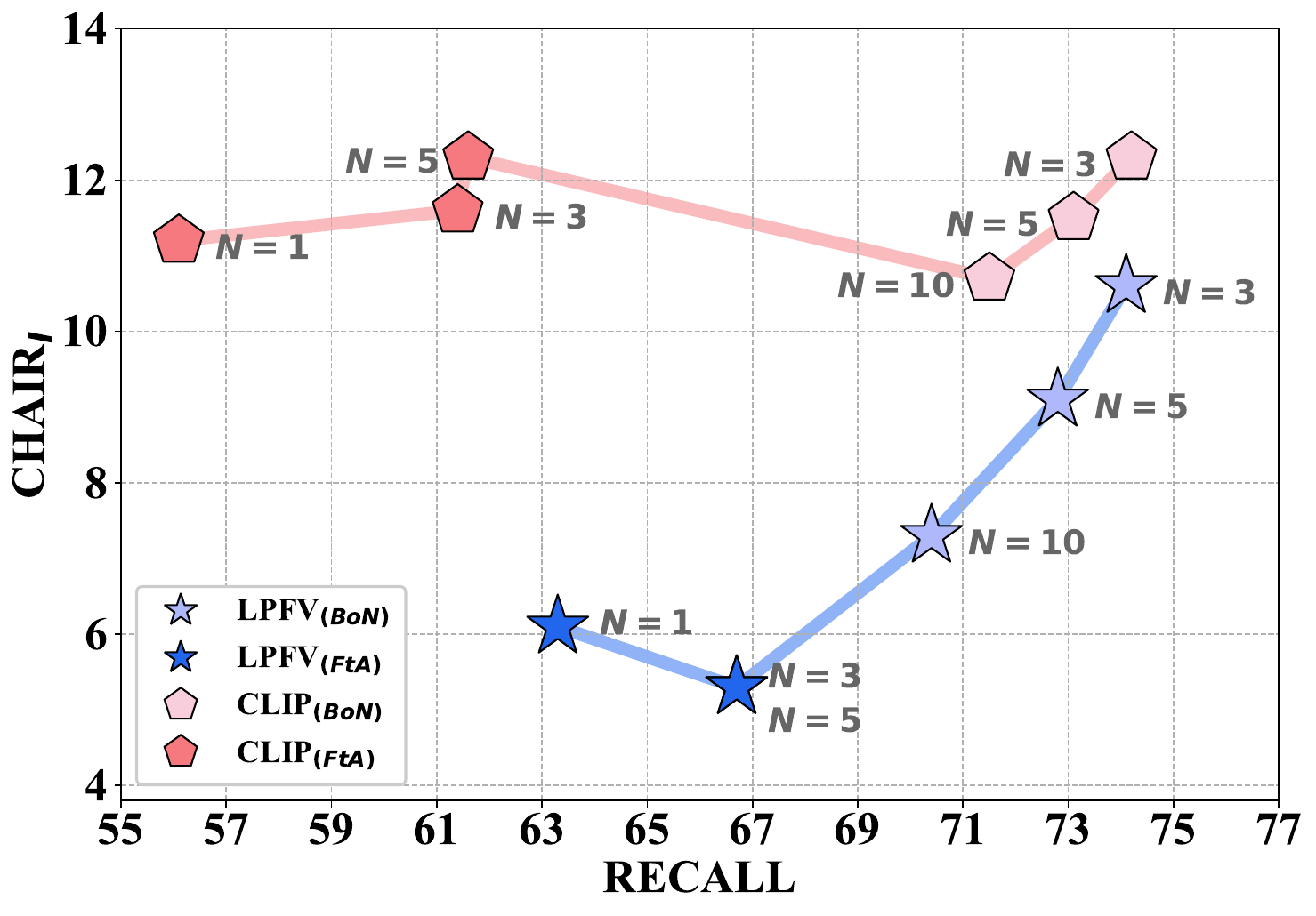} 
    \caption{$\text{CHAIR}_I$ vs. Recall performance, with CLIPScore or LPFV as object verification methods for our framework.}
    \label{fig:verifier}
    \vspace{-0.3em}
\end{figure}

To highlight the importance of the verification in our self-validation framework, we compared two methods for the object verification step: object-level CLIPScore and our LPFV. With $\alpha$ fixed at 0.05, we measured their respective $\text{CHAIR}_I$ and recall scores across various $N$ values. As shown in Figure~\ref{fig:verifier}, CLIPScore performs worse on both BoN and FtA, reducing hallucinations less effectively while also preserving less recall. This performance gap is explained by its weaker discrimination capability; CLIPScore achieves an AUROC of only 0.68, in contrast to 0.85 for LPFV. This result confirms that the accuracy of the verification method is critical to our framework's overall performance.

\section{Reproduction Details of Methods}
\subsection{Reproduction Details of Methods}
\label{app:repro}
We present the results of the compared methods on LLaVA-v1.5-7B in the main experiment.
The hyperparameters for these methods are reported in
Table~\ref{tab:hyperparameter_vcd},
Table~\ref{tab:hyperparameter_cgd},
Table~\ref{tab:hyperparameter_halc},
Table~\ref{tab:hyperparameter_lessismore},
Table~\ref{tab:hyperparameter_deco} and
Table~\ref{tab:hyperparameter_nullu},
respectively. For each method, we follow the official implementation and use the configurations from their respective repositories to reproduce.

\begin{table}[H]
\centering
\begin{tabular}{l|c}
\hline
\textbf{Parameters} & \textbf{Value} \\ \hline
$\alpha$ & $1$  \\ \hline
$\beta$ & $0.1$  \\ \hline
Noise Step & $500$  \\ \hline
\end{tabular}
\caption{VCD Hyperparameter Settings}
\label{tab:hyperparameter_vcd}
\end{table}
\begin{table}[H]
\centering
\begin{tabular}{l|c}
\hline
\textbf{Parameters} & \textbf{Value} \\ \hline
$\alpha$ & $0.01$ \\ \hline
$\beta$ & $1$  \\ \hline
Beam Size & $1$  \\ \hline
Return Sequences Nums $m$ & $3$  \\ \hline
Max Candidate Nums $n$ & $3$  \\ \hline
\end{tabular}
\caption{CGD Hyperparameter Settings}
\label{tab:hyperparameter_cgd}
\end{table}
\begin{table}[H]
\centering
\begin{tabular}{l|c}
\hline
\textbf{Parameters} & \textbf{Value} \\ \hline
Beam Size & $2$  \\ \hline
Candidate Nums $k$ & $2$  \\ \hline
Box Threshold & $0.45$  \\ \hline
Expand Ratio $\lambda$ & $0.6$  \\ \hline
\end{tabular}
\caption{HALC Hyperparameter Settings}
\label{tab:hyperparameter_halc}
\end{table}
\begin{table}[H]
\centering
\begin{tabular}{l|c}
\hline
\textbf{Parameters} & \textbf{Value} \\ \hline
Lora $\alpha$ & $256$  \\ \hline
Lora $r$ & $128$  \\ \hline
Batch Size & $64$\\ \hline
Epoch & $1$\\ \hline
Learning Rate & $2e^{-4}$\\ \hline
Scheduler & cosine\\ \hline
\end{tabular}
\caption{Less Hyperparameter Settings}
\label{tab:hyperparameter_lessismore}
\end{table}
\begin{table}[H]
\centering
\begin{tabular}{l|c}
\hline
\textbf{Parameters} & \textbf{Value} \\ \hline
$\alpha$ & $0.6$ \\ \hline
Beam Size & $1$  \\ \hline
Start Layer & $20$  \\ \hline
End Layer & $28$  \\ \hline
\end{tabular}
\caption{DeCo Hyperparameter Settings}
\label{tab:hyperparameter_deco}
\end{table}
\begin{table}[H]
\centering
\begin{tabular}{l|c}
\hline
\textbf{Parameters} & \textbf{Value} \\ \hline
$\alpha$ & $1$  \\ \hline
Top Ranks $k$ & $4$  \\ \hline
Lowest Layer & $16$  \\ \hline
Highest Layer & $32$  \\ \hline
\end{tabular}
\caption{HALC Hyperparameter Settings}
\label{tab:hyperparameter_nullu}
\end{table}

\subsection{Object Extraction}
\begin{figure*}[t]
   \begin{center}
   \includegraphics[width=0.8\textwidth]{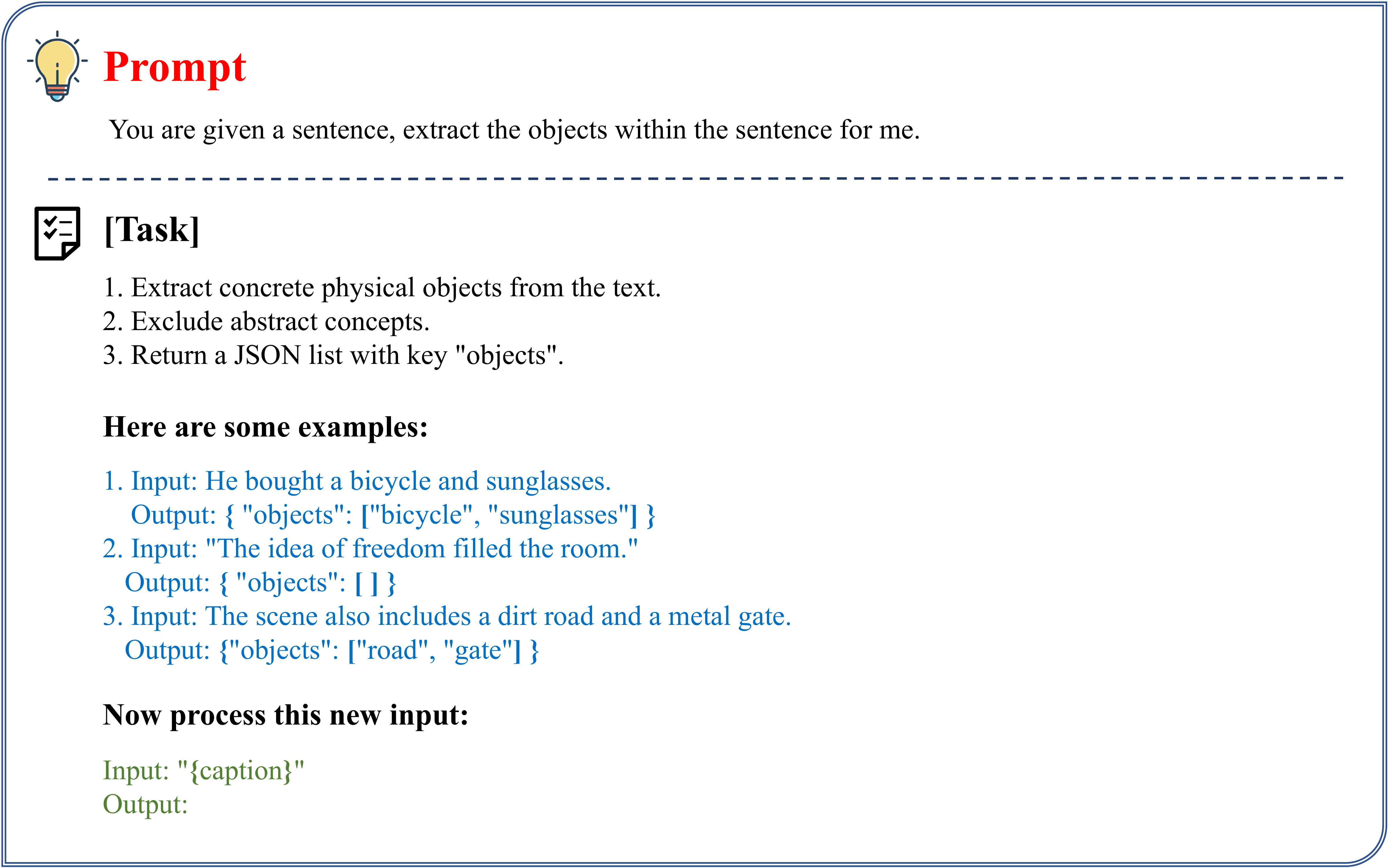}
   \end{center}
   \caption{
     Prompt template for object extraction.
   }
   \label{fig:object-extraction}
\end{figure*}

The template for object extraction by LVLM itself is illustrated in Figure~\ref{fig:object-extraction}.

\subsection{GPT-Assisted Evaluation Template}
\label{app:gpt-eval}
\begin{figure*}[ht]
    \begin{center}
        \includegraphics[width=0.8\linewidth]{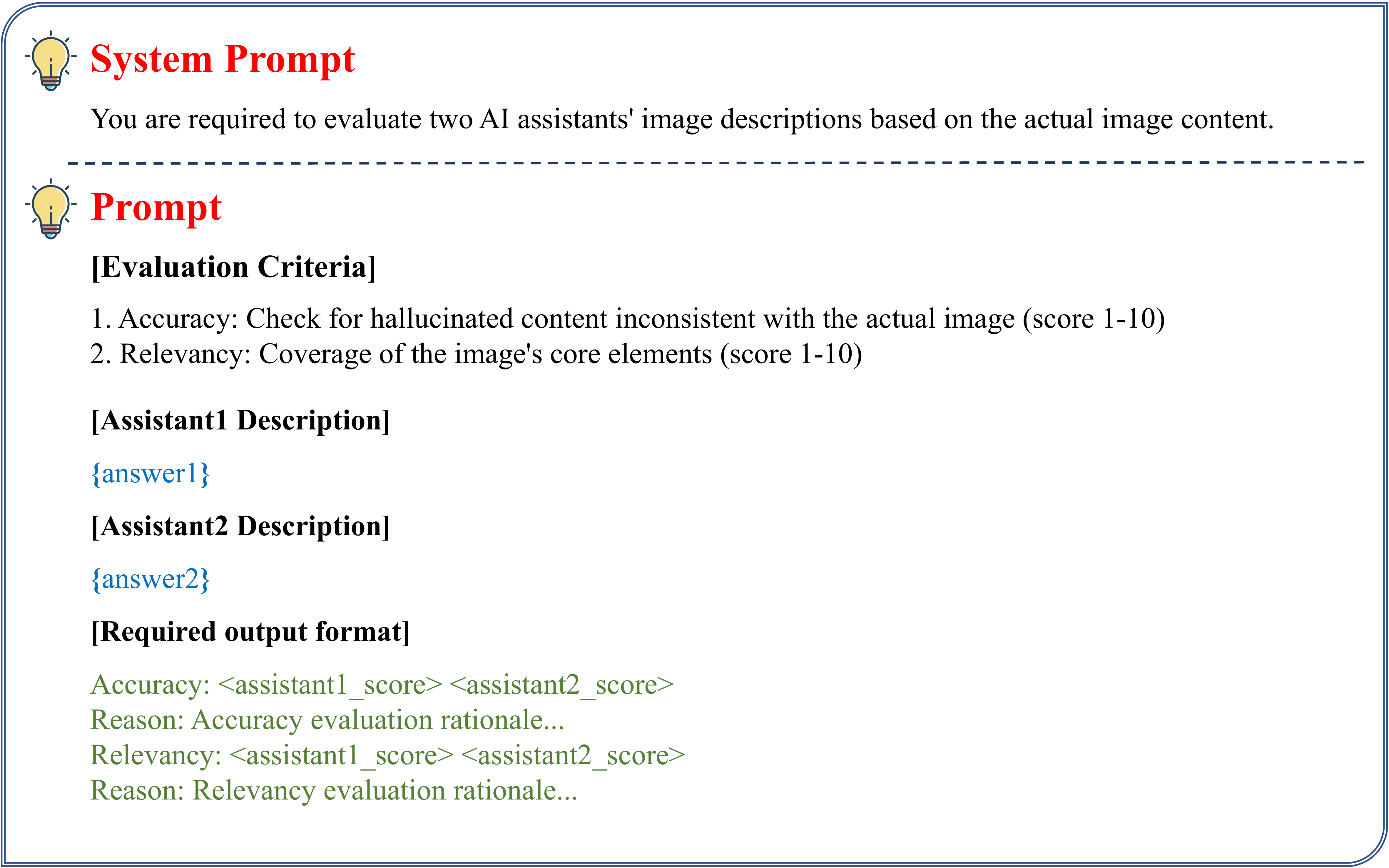}
    \end{center}
    \vspace{-8pt}
\caption{Prompt Template for GPT-Assisted Evaluation.}
\label{fig:gpt-eval}
\end{figure*}

The template for the GPT-assisted evaluation is illustrated in Figure~\ref{fig:gpt-eval}. 
To mitigate position bias (e.g., GPT favoring answer1), we alternate the order so the base caption and each method’s caption swap places as answer1 and answer2. The base model’s final score is the average of its ratings across all pairings.

\section{Qualitative Results}
\begin{figure*}[ht]
    \begin{center}
        \includegraphics[width=0.9\linewidth]{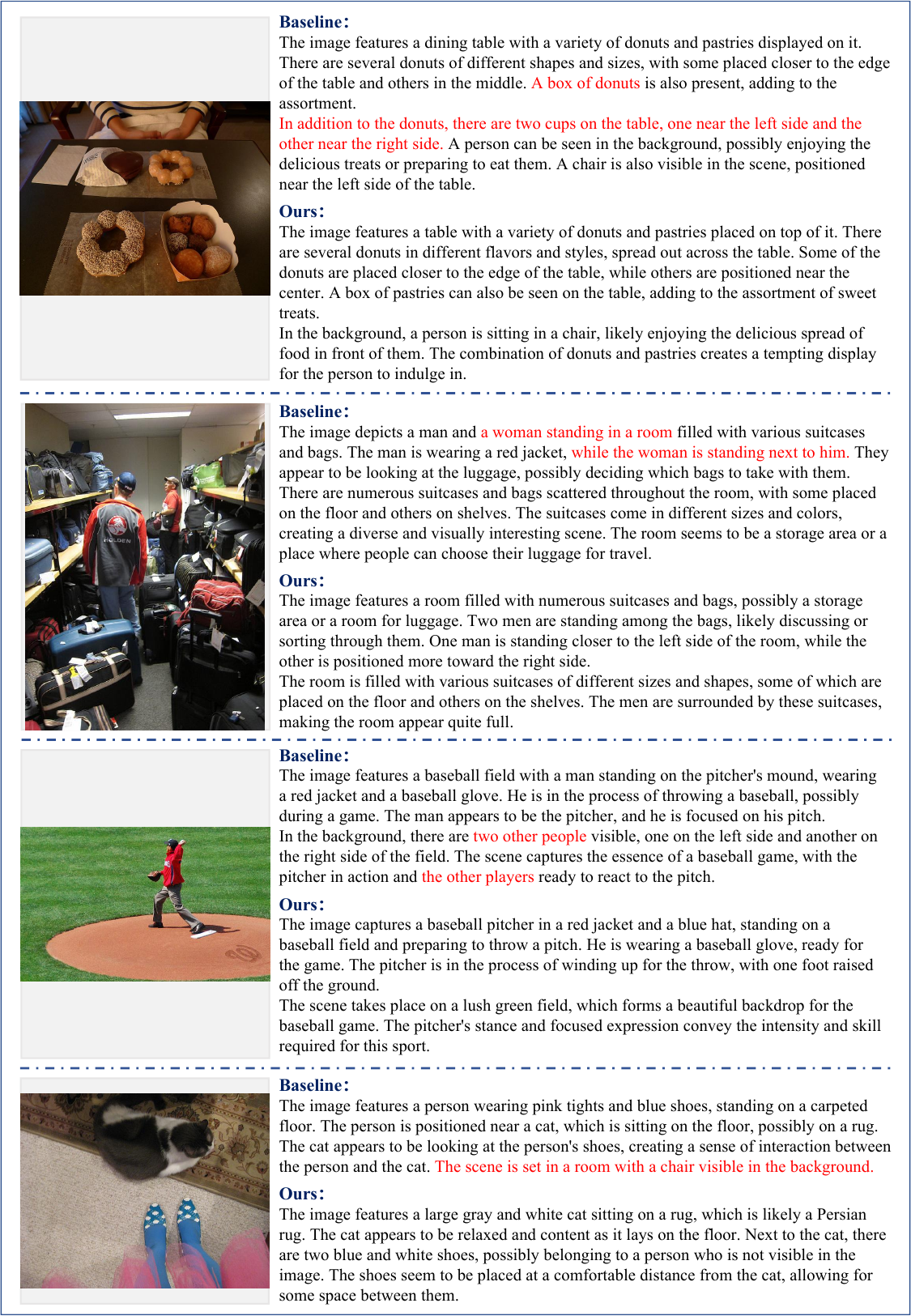}
    \end{center}
    \vspace{-8pt}
\caption{Qualitative results of the LLaVA-1.5-7B model (\textbf{Baseline}) and ours with Best-of-N Selection (\textbf{Ours}).}
\label{fig:quali-bon}
\end{figure*}

\begin{figure*}[ht]
    \begin{center}
        \includegraphics[width=0.9\linewidth]{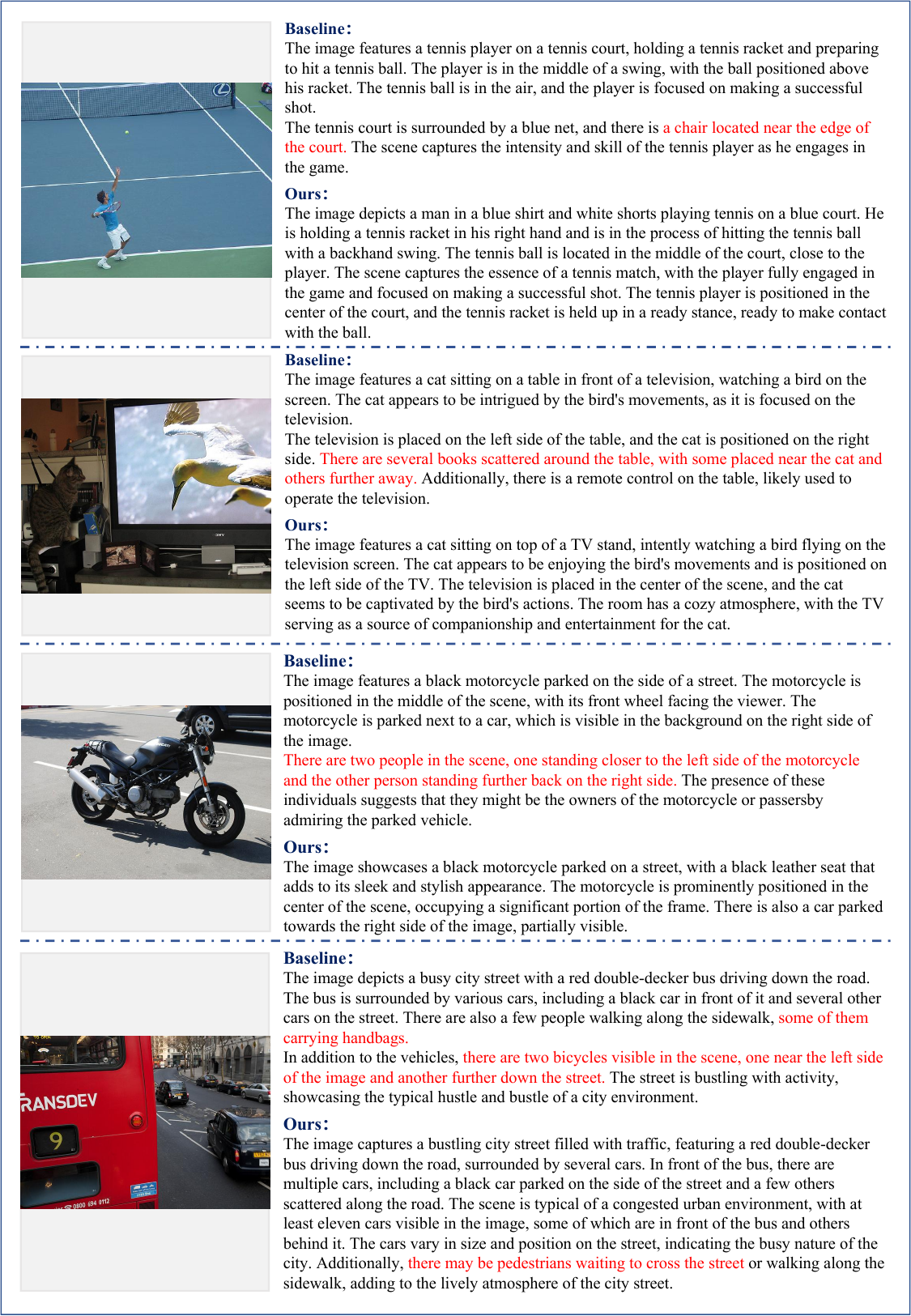}
    \end{center}
    \vspace{-8pt}
\caption{Qualitative results of the LLaVA-1.5-7B model (\textbf{Baseline}) and ours with Filter-then-Aggregate (\textbf{Ours}).}
\label{fig:quali-fta}
\end{figure*}

We provide qualitative demonstrations of our framework's two strategies applied to LLaVA-v1.5-7B. Figure~\ref{fig:quali-bon} compares the Best-of-N Selection strategy against the baseline, while Figure~\ref{fig:quali-fta} presents the same comparison for the Filter-then-Aggregate strategy. In both comparisons, our framework clearly suppresses the object hallucinations, highlighting the effectiveness of both strategies.

\clearpage

\end{document}